\documentclass[12pt]{article}

\usepackage[english]{babel}
\usepackage[utf8]{inputenc}
\usepackage{graphicx}
\usepackage{caption}
\usepackage{subcaption}
\usepackage{dsfont}
\usepackage{verbatim}
\usepackage[colorinlistoftodos]{todonotes}
\usepackage{authblk}
  
\usepackage[tbtags]{amsmath}
\usepackage{amsfonts}
\usepackage{dsfont}
\usepackage{bm}
\usepackage{url}
  
\usepackage{mathtools}
\usepackage{a4wide}

\usepackage[linesnumbered]{algorithm2e}
\usepackage{algorithmic}
\usepackage{soul}
\usepackage{placeins}
\usepackage{float}

\usepackage[symbol]{footmisc}

\usepackage{xr}
\usepackage{lineno}

\makeatletter
\DeclareFontEncoding{LS1}{}{}
\DeclareFontSubstitution{LS1}{stix}{m}{n}
\DeclareMathAlphabet{\mathscr}{LS1}{stixscr}{m}{n}
\makeatother

\usepackage[sort&compress]{natbib}

\title{Recognizing Images with at most one Spike per Neuron}

\author
{Christoph Stöckl$^{1}$, Wolfgang Maass$^{1,\ast}$}

\begin{document}

\maketitle
\newenvironment{affiliations}{%
    \setcounter{enumi}{1}%
    \setlength{\parindent}{0in}%
    \slshape\sloppy%
    \begin{list}{\upshape$^{\arabic{enumi}}$}{%
        \usecounter{enumi}%
        \setlength{\leftmargin}{0in}%
        \setlength{\topsep}{0in}%
        \setlength{\labelsep}{0in}%
        \setlength{\labelwidth}{0in}%
        \setlength{\listparindent}{0in}%
        \setlength{\itemsep}{0ex}%
        \setlength{\parsep}{0in}%
        }
    }{\end{list}\par\vspace{12pt}}

\begin{affiliations}
\item {Institute of Theoretical Computer Science, Graz University of Technology, \\
Inffeldgasse 16b, Graz, Austria \\
$^\ast$ To whom correspondence should be addressed; E-mail: maass@igi.tugraz.at.}
\end{affiliations}
\begin{abstract}
In order to port the performance of trained artificial neural networks (ANNs) to spiking neural
networks (SNNs), which can be implemented in neuromorphic hardware with a drastically reduced energy
consumption, an efficient ANN to SNN conversion is needed. Previous conversion schemes focused on the 
representation of the analog output of a rectified linear (ReLU) gate in the ANN by the firing rate of a spiking neuron.
But this is not possible for other commonly used ANN gates, and it reduces the throughput even for ReLU gates. 
We introduce a new conversion method where a gate in the ANN, which can basically be of any type, is emulated by a small circuit of spiking neurons, with At Most One Spike (AMOS) per neuron. We show that this AMOS conversion
improves the accuracy of SNNs for ImageNet from 74.60\% to 80.97\%,
thereby bringing it within reach of the best available ANN accuracy (85.0\%). The Top5 accuracy of SNNs is raised to 95.82\%, getting even closer to the best Top5 performance of 97.2\% for ANNs.
In addition, AMOS conversion improves
latency and throughput of spike-based image classification by several orders of magnitude. 
Hence these results suggest that SNNs provide a viable direction for developing highly energy efficient hardware for AI that combines high performance with versatility of applications.
\end{abstract}

\section*{Introduction}

Spiking neural networks (SNNs) are among the leading candidates to solve one of the major impediments
of more widespread uses of modern AI: The energy consumption of the very large artificial neural
networks (ANNs) that are needed. These ANNs need to be large, since they need to have a sufficiently 
large number of parameters in order to absorb enough information from the huge data sets on which they 
are trained, such as the 1.2 million images of ImageNet2012.
Inference on these large ANNs is power hungry \\
\citep{Garcia-Martin2019}, which impedes their deployment in mobile devices or autonomous vehicles.
Spiking neurons have been in the focus of recent work on novel computing hardware for AI with a 
drastically reduced energy budget, because the giant SNN in the brain –consisting of about 100 billion 
neurons-- consumes just 20W \citep{LingJ2001}. Most spiking neuron models that are considered in neuromorphic
hardware are in fact inspired by neurons in the brain. Their output is a train of stereotypical electrical 
pulses --called spikes. Hence their output is very different from the analog numbers that an ANN neuron
produces as output.

But whereas large ANNs, trained with ever more sophisticated learning algorithms on 
giant data sets, approach --and sometimes exceed-- human performance in several categories of intelligence,
the performance of SNNs is lagging behind. One strategy for closing this gap is to design an ANN-to-SNN 
conversion that enables us to port the performance of a trained ANN into an SNN. The most common –and so 
far best performing—conversion method was based on the idea of (firing-) rate coding, where the analog
output of an ANN unit is emulated by the firing rate of a spiking neuron \citep{Rueckauer2017}.
This method can readily be used for ANN units that are based on the ReLU (rectified linear) activation 
function. It has produced impressive results for professional benchmark tasks such as ImageNet,
but a significant gap to the accuracy, latency, and throughput of ANN solutions has thwarted its
practical application. Problems with the timing and precision of resulting firing rates on higher 
levels of the resulting SNNs have been cited as possible reasons for the loss in accuracy of the SNN. 
In addition, the transmission of an analog value through a firing rate requires a fairly large number
of time steps –typically in the order of 100, which reduces both latency and throughput for inference. 
An additional impediment for a rate-based ANN-to-SNN conversion emerged more recently in the form of
better performing ANNs such as EfficientNet \citep{Tan2019}. They employ the Swish function as 
activation function, defined by $x\cdot \text{sigmoid}(x)$,  
which contains more complex non  
linearities than the ReLU function. Furthermore, the Swish function also produces negative values that 
can not be readily encoded by the -- necessarily non-negative -- firing rate of a spiking neuron.  

We introduce a new ANN-to-SNN conversion method that we call AMOS conversion because it requires 
At-Most-One-Spike (AMOS) per neuron. This method is obviously very different from rate-based conversions,
and structurally more similar to temporal coding, where the delay of a single spike is used to encode an 
analog value. 
However temporal coding has turned out to be difficult to implement in a noise-robust
and efficient manner in neuromorphic hardware. This arises from the difficulty to implement delays with 
sufficiently high precision without sacrificing latency or throughput of the SNN, and the difficulty to 
design spiking neurons that can efficiently process such temporal code  \citep{maass1998}, 
\citep{Thorpe2001}, \citep{Rueckauer2017}, \citep{Kheradpisheh2019}. In contrast, AMOS coding 
requires just on the order of log N different delays for transmitting integers between 1 to N. 
Furthermore, these delays can be arranged in a data-independent manner that supports pipelining, so that a new image can be processed by the SNN at every time step.

We show that even 
the simplest type of spiking neuron, the McCulloch Pitts neuron or threshold gate
\citep{McCullochPitts43}
can efficiently compute with AMOS codes.
Thus no temporal integration of information is needed for the spiking neuron model in
order to efficiently emulate inference by ANNs.
This simple version of a spiking neuron had previously already been used for image classification by SNNs
in hardware \citep{Esser2016}.
We will describe in the first subsection of Results the design of an AMOS unit that replaces the gate 
of an ANN –with basically any activation function—in this new ANN-to-SNN conversion. We then show that 
this conversion produces an SNN that carries out inference for classifying images from the full ImageNet2012
dataset with drastically improved accuracy, latency, and throughput. Whereas the design of the
AMOS unit for the conversion of ANN-neurons with the Swich function requires training of its parameters,
one can design an AMOS unit for the special case of the ReLU activation function explicitly: It reduces
in this special case to an analog-to-digital conversion via binary coding. This will  be made explicit
in the last subsection of Results.

\section{Results}
\subsection{Architecture of the AMOS unit}
We present in Fig. 1B the architecture of an AMOS unit --consisting of K spiking neurons-- 
that approximates a generic ANN gate with activation function $f$ shown in Fig. 1A.
Besides fixed delays (shown in green) the AMOS unit contains weight coefficients
$c_1$, ..., $c_K$,  $d_1$, ..., $d_K$, $h_{ij}\ \text{for}\ i,j \in \{1, \ldots,K\}$, and thresholds 
$T_1$, ..., $T_K$ (shown in blue).

The case that the activation function $f$ of an ANN gate outputs positive and negative numbers 
is of particular importance
in view of the Swish function (see Fig. \ref{amos-swish}) that was introduced in 
\citep{Ramachandran2017} and used as activation function in 
EfficientNet \citep{Tan2019}.  It is defined by

\begin{equation}
\text{Swish}(x) = x \cdot \frac{1}{1 + e^{-x}}.
 \label{swish}
\end{equation}

\begin{figure}[]
\centering
 \includegraphics[scale=0.6]{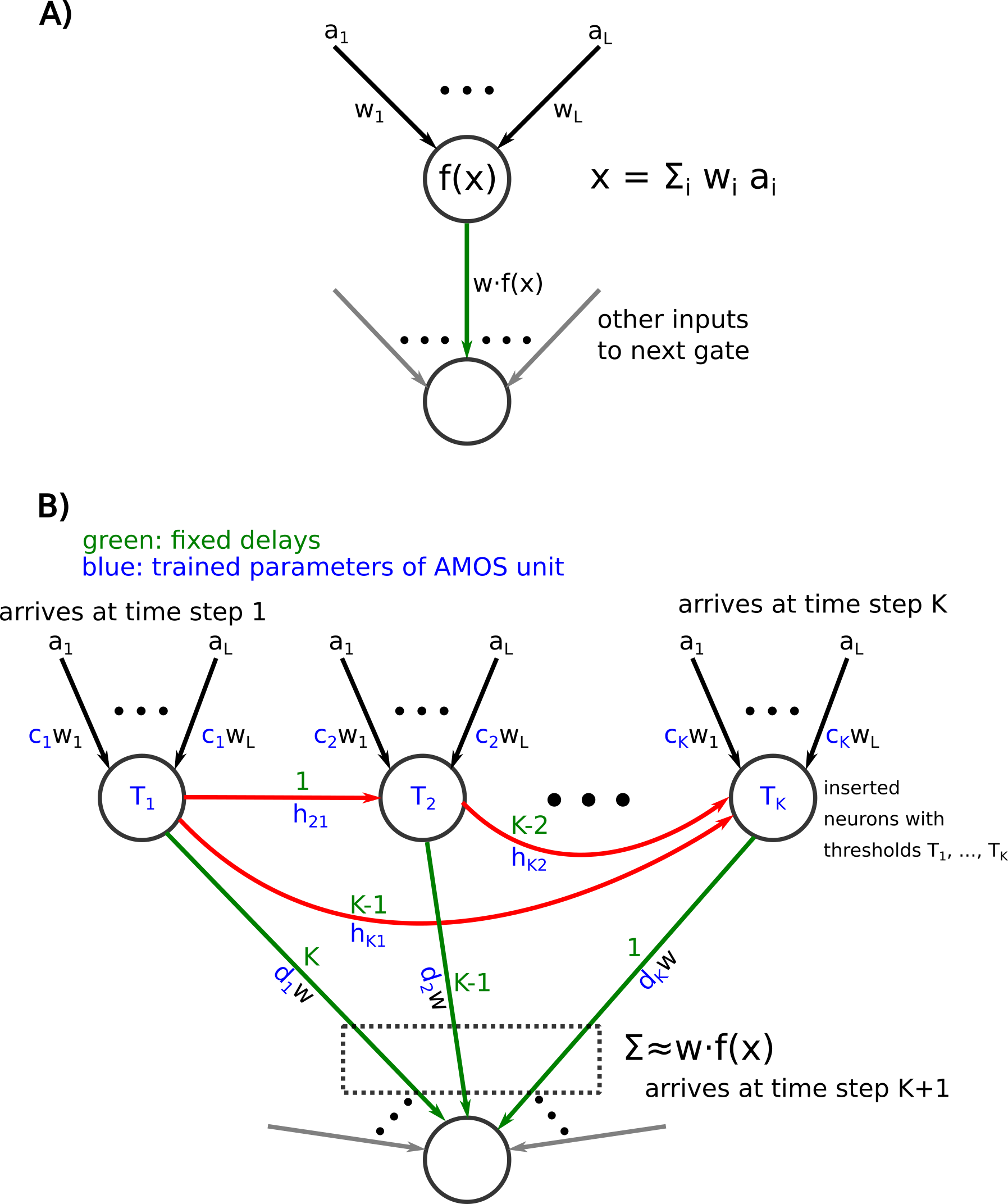}
 \vspace*{10mm}
\caption{\textit{Architecture of an AMOS unit, consisting of $K$ spiking neurons. \\
\textbf{A)} An ANN gate with activation function $f$ that is to be emulated by the AMOS unit. 
\textbf{B)} The AMOS unit receives the same inputs $a_1, \ldots, a_L$, duplicated $K$ times.
It outputs after $K+1$ time steps an approximation of the value  $wf(x)$ which the ANN gate sends 
to subsequent gates.}}
 \label{amos-plt}
\end{figure} 
Neuron $i$ in the AMOS unit outputs
\begin{equation}
 z_i = \Theta(c_i\cdot x - H_i - T_i ),
 \label{spike-fn}
\end{equation}
where $\Theta$ is the Heaviside activation function defined by
\begin{equation}
 \Theta(x)  =
    \begin{cases}
      0, & \text{if}\ x < 0 \\
      1, & \text{if}\ x \geq 0,
    \end{cases}
\end{equation}
the coefficient $c_i$ and the threshold $T_i$ are trained parameters,
and $H_i$ is an inhibitory input defined by 
\begin{equation}
H_i = \sum_{j=1}^{i-1} h_{ij} z_{j}
\end{equation}
with trained negative weights $h_{ij}$. 
The output $y$ of the AMOS unit, which is fed into subsequent AMOS units
that emulate subsequent gates to which the ANN that it emulates is connected 
(only one subsequent gate is shown for simplicity in Fig. \ref{amos-plt}), can be written as
\begin{equation}
 y = \sum_{i=1}^{K} wd_i z_j, 
\end{equation}
where the $d_i$ are additional trained weights of the AMOS unit,
and $w$ is the weight of the corresponding connection from the ANN gate 
in Fig. 1A to the next ANN gate. Thus the computational function of the entire AMOS unit
can be expressed as
\begin{equation}
 \text{AMOS}(x) = y =  
 \sum_{i=1}^{K} w \cdot d_i \cdot \Theta(c_i \cdot x - H_i - T_i). 
\end{equation}

\begin{figure}[]
\centering
 \includegraphics[scale=0.7]{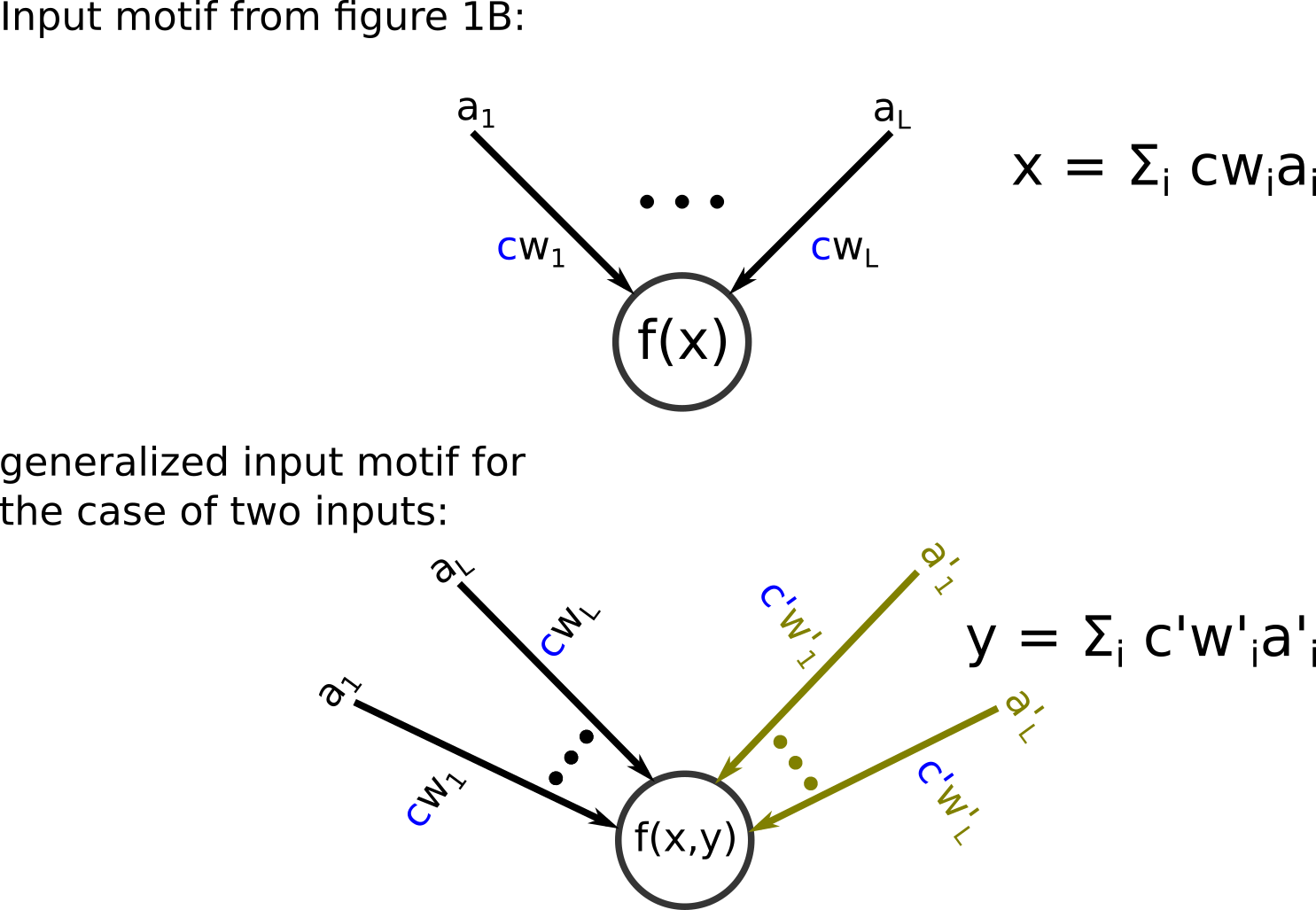}
 \vspace*{10mm}
\caption{\textit{AMOS approximation of a function $f(x,y)$ with two input values $x,y$, e.g. $x \cdot y$ }}
\label{one-to-many}
\end{figure}

For the case of functions $f(x,y)$ with two input values $x$ and $y$, one just needs to 
replace the motif of Fig.~\ref{amos-plt}A in Fig.~\ref{amos-plt} by the motif indicated 
in Fig.~\ref{one-to-many}B. 

 
This changes equation \ref{spike-fn} to:
\begin{equation}
z_i = \Theta(c_i\cdot x + c_i' \cdot x' - H_i - T_i )
\end{equation}

\subsection{Approximation of some common ANN gates by AMOS units}
In the case of the ReLU activation function no training of the parameters of the AMOS unit from Fig. 1B
is needed. If one defines its thresholds $T_j$ by $2^{K - j}$ and weights $d_j, c_j, h_{ij}$ 
by $c_j = 1, d_j=2^{K-j}, h_{ij} = 2^{K-j}$, an AMOS unit with $K$ neurons
produces an approximation $\text{ReLU}(x)$ of the activation function ReLU(x) that deviates for x from $0$ by at 
most $\alpha2^{-K}$ for an approximation in the interval from $-\infty$ to $\alpha$, for any $\alpha \in \mathbb{R}$. The resulting approximation is plotted in Fig. 3.

For the case of other gate functions $f$ we train the additional weights and thresholds of 
the AMOS unit through backpropagation, using a triangle-shaped pseudo-derivative in place of 
the non-existing derivative of the Heaviside function.
Results of such approximations are plotted in Fig.~\ref{amos-sigmoid} - \ref{fig6} for the case 
of the sigmoid function, Swish function, and multiplication.


Note that the AMOS unit can be used in a pipelined manner, processing $K$ different
inputs $x$ in its $K$ time steps. Hence the resulting SNN can be used in a pipelined manner,
processing a new network input at each time step.
Hence its throughput is much better than that of SNNs that result from rate-based ANN-to-SNN 
conversions, such as for example \citep{Rueckauer2017, Sengupta2019}.

The number of neurons in the network is increased through the AMOS conversion by some factor. 
However a hardware implementation can reuse AMOS units for multiple time steps
(since all AMOS units have the same internal parameters), thereby reducing the required size
of the network at the cost of a corresponding reduction of the throughput.

\begin{figure}[H]
\centering
 \includegraphics[scale=0.7]{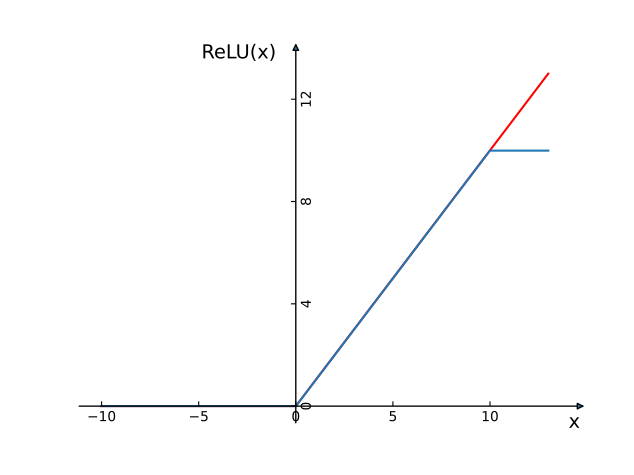}
\caption{\textit{AMOS approximation of the ReLu function, $K=10$, \\
(red: target function, blue: AMOS approximation)}}
\label{amos-relu}
\end{figure}

\begin{figure}[H]
\centering
 \includegraphics[scale=0.7]{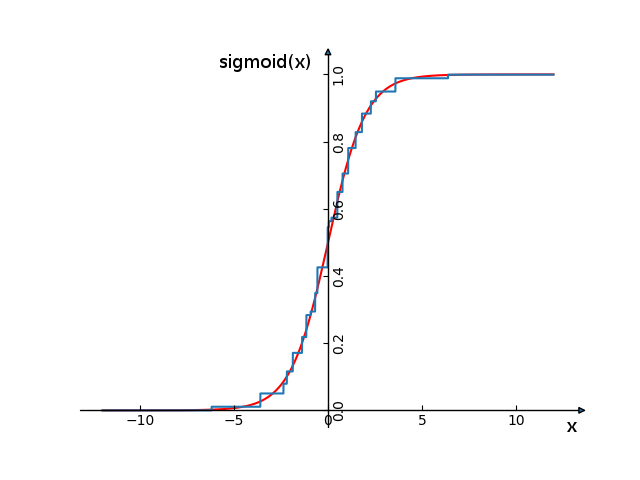}
\caption{\textit{AMOS approximation of the sigmoid function, $K=8$ \\
(red: target function, blue: AMOS approximation)}}
\label{amos-sigmoid}
\end{figure} 

\begin{figure}[H]
	\centering
	\includegraphics[scale=0.6]{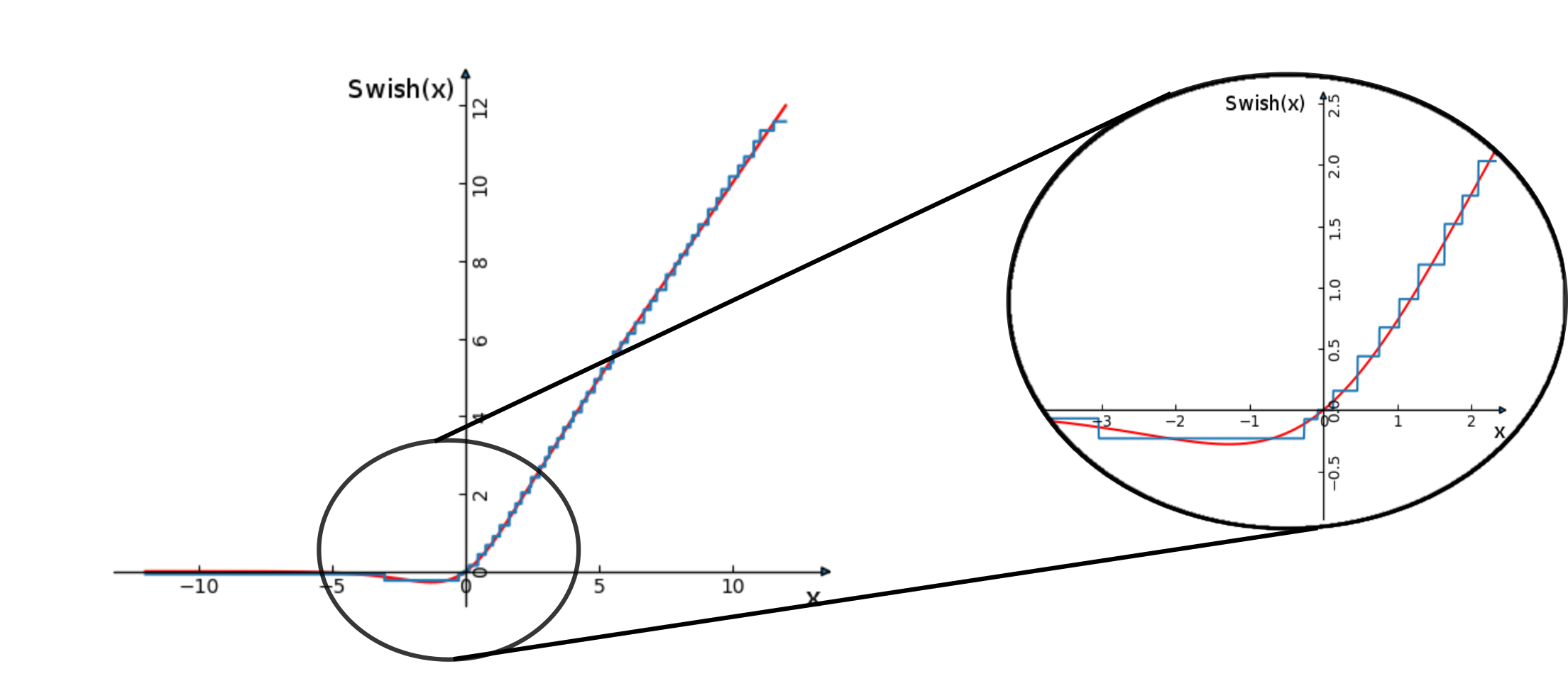}
	\caption{\textit{AMOS approximation of the Swish function, $K=12$, \\
			(red: target function, blue: AMOS approximation)}}
	\label{amos-swish}
\end{figure} 
\begin{figure}[H]
	\centering
	\includegraphics[scale=0.55]{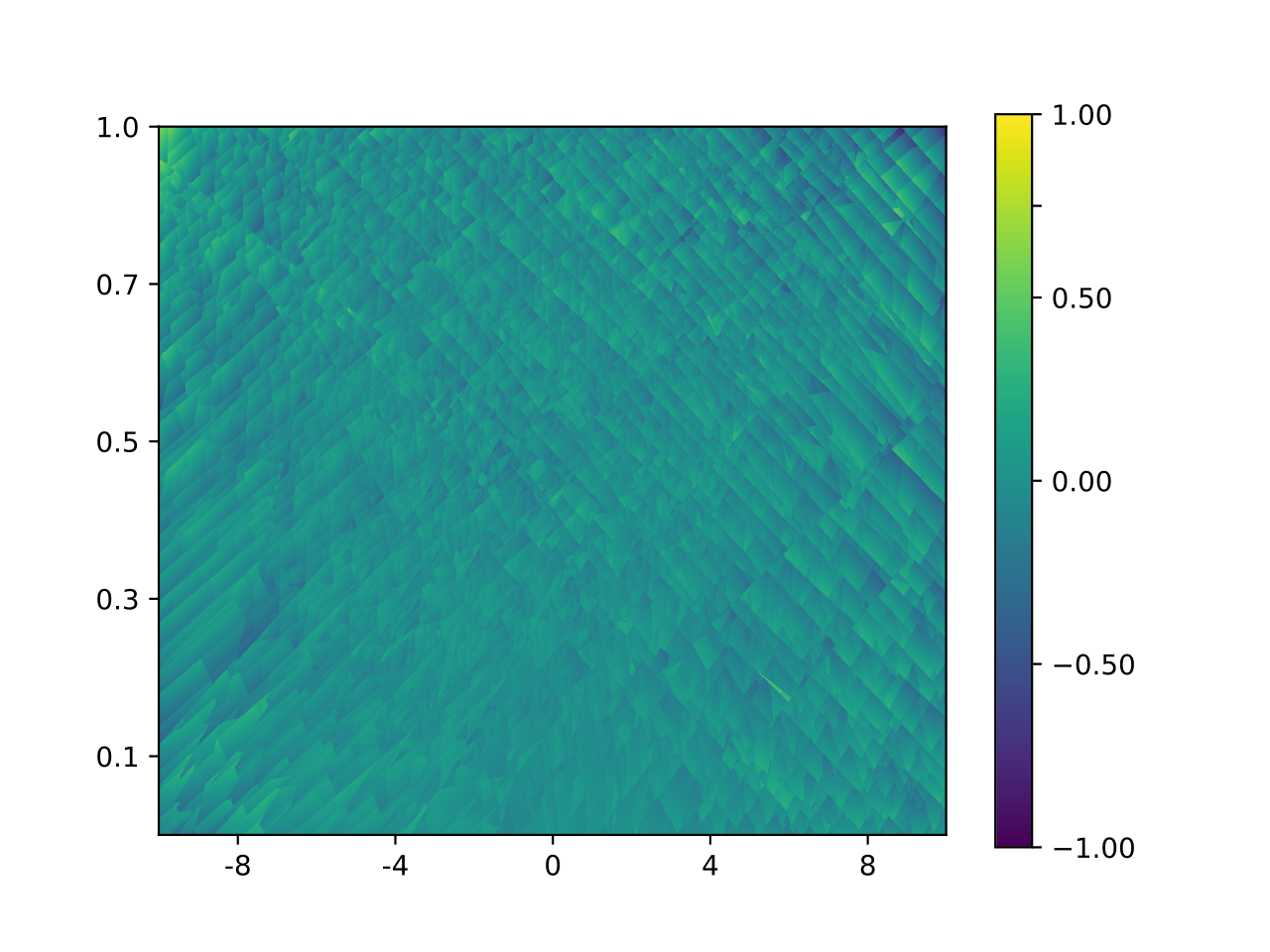}
	\caption{\textit{Approximation error of the trained AMOS unit for multiplication with $K = 40$ (MSF = 0.0102).}}
	\label{fig6}
\end{figure} 

\subsection{Application of the AMOS conversion to image classification with SNNs}
The ImageNet data set \citep{Russakovsky2015} has become the most popular benchmark
for image classification in machine learning (we are using here the ImageNet2012 version).
This data set consists of $1.281.167$ training images and $50.000$ test images 
(both RGB images of different sized), that are labeled by 1000 different categories. 
Classifying imaged from ImageNet is a nontrivial task even for a human, since this data
set contains for example 59 categories for birds of different species and gender \citep{van2015building}. 
This may explain why a relaxed performance measurement, where one records whether the 
target class is among the top 5 classifications that are proposed by the neural network ("Top5"),
is typically much higher.

A new record in ANN performance for the classification of images from the ImageNet 2012 dataset was achieved by \citep{Tan2019}.
They introduced a new ANN family called EfficientNet, that achieved 84.4\%.
With a modified training procedure it can achieve 85\% 
accuracy \citep{Cubuk2019}.
The parameters of the trained network are publicly available\footnote{https://storage.googleapis.com/cloud-tpu-checkpoints/efficientnet/randaug/efficientnet-b7-randaug.tar.gz}.

This accuracy was achieved by the EfficientNet-B7
model that has 66M parameters. Besides a new scaling method for
balancing network depth, width, and image resolution, they also introduced the Swish function as 
activation function, instead of ReLU.
The Swish function emerged from automatic search for better performing activation functions
\citep{Ramachandran2017}. 
Other nonlinearities that are used in EfficientNet are sigmoids and multiplication.
These occur in "Squeeze and Excitation Layers" \citep{Hu2018} of EfficientNet, see Fig. 7. 
The main building block of the EfficientNet architecture is the mobile inverted bottleneck \citep{Sandler2018},
which uses depthwise separable convolutions. 
This type of convolutional layer uses neurons with linear activation functions.
Although it would certainly be possible to approximate linear functions using AMOS conversion, 
we 
simply collapsed linear layers 
into 
the generation of the weighted sums that form the inputs to the next layers.

We evaluated the classification performance of the SNN that results from an application of the 
AMOS conversion described in the previous section to EfficientNet-B7. 
The resulting SNN 
achieved a classification 
performance of $80.97\%$, and $95.82\%$ for Top5; see Table~\ref{tab:my-table-1}. 
The values for $K$ used in the AMOS conversion are listed in Table \ref{K-param-num} .

\begin{table}[H]
\centering
\begin{tabular}{|l|l|l|l|l|l|l|}
\hline
Model & \# params & ANN & SNN & \# layers & \# neurons & \# spikes \\ \hline
EfficientNet-B7 & 66M& 85\% (97.2\%) & 80.97\% (95.82\%) & 4605 & 3110M & 1799M \\ \hline
ResNet50 & 26M & 75.22\% (92.4\%) & 75.10\% (92.36\%) & 500 & 96M & 14M \\ \hline
\end{tabular}
\caption{\textit{Performance results for ANNs and the SNNs, that result from their 
AMOS conversion on ImageNet. Top5 accuracy is given in parentheses. 
 \#layers and \#neurons
refer to the SNN version of the network. Spike numbers refer to inference for a sample test image.}}
\label{tab:my-table-1}
\end{table}

Note that the resulting SNNs have an optimal throughput, since they can classify a new image 
at each time step. SNNs that result from a rate-based ANN conversion need up to 3000 time steps 
for reaching maximum accuracy, hence their throughput is by a corresponding factor smaller. 
The number of parameters is increased by the AMOS conversion only by a small additive
term (see Table \ref{K-param-num}), since all AMOS units of the same type use the same parameters.

The accuracy of 75.22\%  for the ANN version of ResNet50 in Table \ref{tab:my-table-1} resulted from 
training a variant of ResNet50 where max pooling was replaced by average pooling, using
the hyperparameters given in the TensorFlow repository. 
This accuracy is close to the best published performance of 76\% for ResNet50 ANNs \citep[Table 2]{Tan2019}. 
Apart from max-pooling, ResNets use neither Swish nor sigmoid or multiplication as nonlinearities, just ReLU. 
This explains why the application of the AMOS conversion to ResNet yields SNNs whose Top1 and Top5
performance is almost indistinguishable from the ANN version. Note also that the resulting SNNs
are only sparsely active, an activity regime that enhances the energy efficiency of some hardware 
implementations.

The best previous performance of an SNN on ImageNet was achieved by 
converting an Inception-v3 model \citep{Szegedy2016} with a rate-based 
conversion scheme \citep{Rueckauer2017}.
The reported test accuracy of the resulting SNN was 74.6\%, where 550 time steps were used 
to simulate the model.
Hence already the application of AMOS conversion to ResNet50 improves this result with regard
to accuracy, and especially with regard to throughput. The AMOS conversion of EfficientNet-B7 
yields an additional $5.87\%$ accuracy improvement.

\subsection{Results for the classification of images from the CIFAR10 data set}

The results for the ANN versions of ResNet that are given in Table \ref{tab:my-table} are the outcome of 
training them with the hyperparameters given in the TensorFlow models repository.
They are very close to the best results reported in the literature.
The best ResNet on CIFAR10 is the ResNet110, where a test accuracy of 93.57\% has been reported \citep{He2016}.
Our ResNet50 achieves 92.99\%, which is very close to the performance of the ResNet56 with 93.03\%.

Spiking versions of ResNet20 have already been explored \citep{Sengupta2019}.
Using a rate-based conversion scheme a performance of 87.46\% was reported. 
Compared to these results, AMOS conversion yields a higher accuracy, while also 
using significantly fewer time steps, thereby reducing latency for inference. 
In addition, the throughput is reduced from a rather low value to the theoretically ideal 
value of one image per time step. \\

\begin{table}[]
\centering
\begin{tabular}{|l|l|l|l|l|}
\hline
Model & ANN & SNN & \# neurons & \# spikes \\ \hline
ResNet50 & 92.99\% & 92.42\% & 4.751.369 & 647.245 \\ \hline
ResNet20 & 91.58\% & 91.45\% & 1.884.160 & 261.779 \\ \hline
ResNet14 & 90.49\% & 90.39\% & 1.310.720 & 190.107 \\ \hline
ResNet8 & 87.22\% & 87.05\% & 737.280 & 103.753 \\ \hline
\end{tabular}
\caption{\textit{SNN performances on CIFAR10 that result from AMOS conversions of ResNet models of different
depths. (using $K=10$ in the AMOS units) \#neurons refers to the SNN version. }}
\label{tab:my-table}
\end{table}

\subsection{Trade-off between latency and network size of the SNNs}

It is well-known that the extraction of bits from a weighted sum, as well as multiplication
of binary numbers, can be carried out by threshold circuits --hence also by SNNs-- with a small 
number of layers --typically 2 or 3-- that does not depend on the bit length of the binary numbers involved,
however at the cost of increasing the number of neurons to a low-degree polynomial of this bit length.
A recent summary of such results is provided in section 3 of \citep{parekh2018constant}. 
Hence one can replace the basic architectures of the AMOS units from Fig. 1 and 2 by the more
shallow architectures that employ a larger number of spiking neurons. In order to make the 
resulting circuits applicable to larger input domains, one can train their parameters, similarly 
as for the basic architectures.
Hence there exists a trade-off between the number of layers (latency) and the 
size of the SNN that can be achieved through an AMOS conversion, and several points 
on this trade-off curve can be reached through these modified AMOS conversions. 

\section{Methods}

\subsection{Squeeze and Excitation in EfficientNet}

\begin{figure}[H]
	\centering
	\includegraphics[scale=0.5]{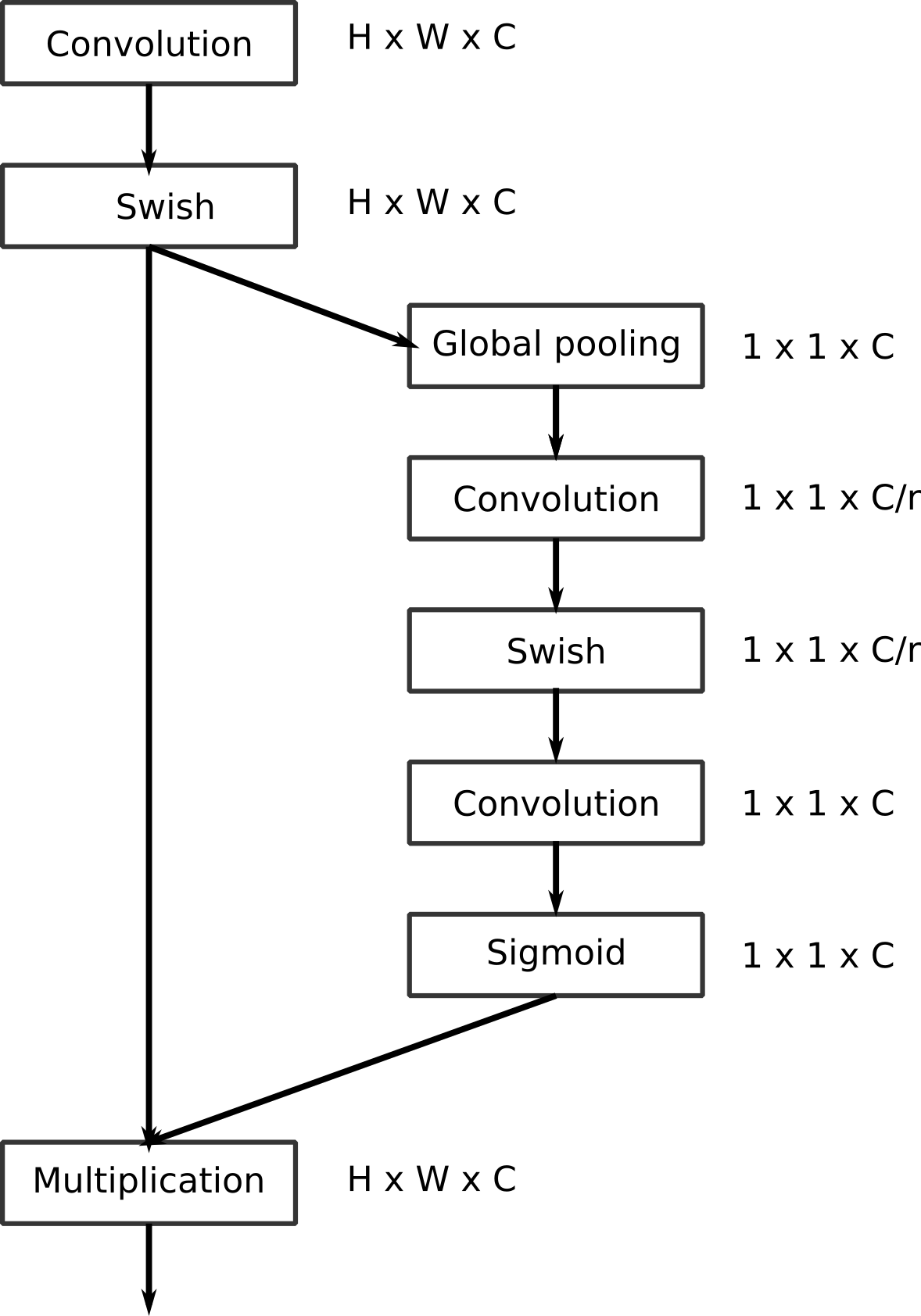}
	\caption{Squeeze and Excitation Layer}
	\label{fig:7_new}
\end{figure} 




\subsection{Number of parameters of AMOS units}



\begin{table}[H]
	\centering
	\begin{tabular}{|l|l|}
		\hline
		Approximation & Number of parameters \\ \hline
		Sigmoid (K=8) & 52 \\ \hline
		ReLU (K= 10) & 75 \\ \hline
		Swish (K=12) & 102 \\ \hline
		Mult (K= 40) & 940 \\ \hline
	\end{tabular}
	\caption{Number of parameters in the AMOS units that are used in this article}
	\label{K-param-num}
\end{table}




\section*{Discussion}

We have introduced a new method for converting ANNs to SNNs. 
Since the resulting SNN uses for inference at most one spike (AMOS) per neuron, 
AMOS conversion can be seen as dual to the familiar rate-based conversion,
since it uses space -- i.e., more neurons -- rather than time for replacing analog values by spikes.
This conversion significantly improves accuracy, latency,
and especially the throughput of the resulting SNN. Furthermore, since AMOS conversion can be applied
to virtually any type of ANN gate, it demonstrates for the first time that SNNs are universal computing
devices from the perspective of modern ANN-based machine learning and AI. For the classification of 
natural images, it raises the accuracy of SNNs for the full ImageNet 2012 dataset
to 80.97\% -- and to 95.82\% for Top5 -- thereby bringing it into the range of the best ANN and human performance. 
This 
was achieved by converting EfficientNets, a recently proposed variant
of ANNs that employ the Swish function as neural activation function, for which a rate-based conversion
to SNNs is impossible. The resulting SNNs that achieve high performance 
are  -- like their
ANN counterparts -- very large. But one can sacrifice some of their accuracy by starting with a smaller
ANN, or by reducing their perfect throughput of one image per step and reuse gates of a smaller SNN with 
online reconfigurable connections and parameters. Note that reuse of neurons 
would be more problematic 
for rate-based SNN computations, since each spiking neuron is occupied there during a fairly large and
somewhat unpredictable number of time steps when emulating a computation step of an ANN gate.
In contrast, AMOS conversion provides a tight bound on the number of time steps during which a spiking 
neuron is occupied. Hence it can also be used for converting recurrently connected ANNs to SNNs. 

Altogether the proposed method for generating highly performant SNNs
offers an opportunity to combine the computationally more efficient and 
functionally more powerful training of ANNs with the superior energy-efficiency
of SNNs for inference. 
Note that one can also use the AMOS-converted SNN as initialization for subsequent direct training 
of the SNN for a more specific task. 
Altogether our results suggest that spike-based hardware may gain an edge in the competition for the development of drastically more energy efficient hardware for AI applications by combining energy efficiency and competitive performance with a versatility that optimized hardware for specific ANNs –such as a specific type of convolutional neural networks— cannot offer.

\section*{Acknowledgements}
We would like to thank Franz Scherr for helpful discussions. This research was partially supported by the Human Brain Project of the European Union (Grant agreement number 785907). 









\bibliographystyle{apalike}
\bibliography{scibib}

\end{document}